\documentclass[10pt,twocolumn,letterpaper]{article}

\usepackage[pagenumbers]{cvpr} 

\newcommand{\name}{ReActWAM}

\usepackage{colortbl}
\definecolor{methodrowblue}{RGB}{232,243,252}

\definecolor{cvprblue}{rgb}{0.21,0.49,0.74}
\usepackage[pagebackref,breaklinks,colorlinks,allcolors=cvprblue]{hyperref}

\def\confName{CVPR}
\def\confYear{2027}

\title{\name: Action-Consistent Caching for Efficient World-Action Models}

\author{Zhaochen Guo \quad Yunke Wang$^{1}$ \quad Siyu Xu$^{1}$ \quad Zijian Wang$^{1}$\\
Chenghao Xia$^{1}$ \quad Chengbin Du$^{1}$ \quad Shinan Liu$^{2}$ \quad Chang Xu$^{1}$\\[0.4em]
$^{1}$The University of Sydney \qquad $^{2}$The University of Hong Kong}

\begin{document}
\raggedbottom
\maketitle
\begin{abstract}
 World-action models (WAMs) often generate future visual states alongside robot actions, introducing substantial computational overhead during inference. During closed-loop execution, consecutive inference cycles often predict similar action sequences, leading to redundant computation. However, reducing this redundancy requires distinguishing consistent actions from changes that demand fresh computation. To address these challenges, we propose ReActWAM:, a training-free framework that combines action-intent-guided prefix skipping with inner-chunk prediction reuse. A coarse-to-fine assessment evaluates motion changes, visual consistency, and action prediction conflicts. For eligible chunks, we draw on first-step differences to adapt the cached latent and network features to the current condition, providing an estimate at a later denoising step that bypasses the redundant prefix. Subsequent inference refines this estimate while reusing stable denoising predictions within the chunk. Experiments across LingBot-VA, FastWAM and the Cosmos family on LIBERO-Long and RoboTwin 2.0 demonstrate $1.41\times$--$5.44\times$ GPU inference speedup, with minimal degradation in task success rates. Real-world experiments on a Franka Research 3 (FR3) robot further achieve $1.46\times$ end-to-end speedup while maintaining success rate.
\end{abstract}

\section{Introduction}
\label{sec:introduction}

World-action models (WAMs) couple action prediction with future-state modeling to support robotic manipulation~\cite{lyu2025dywa,ye2026worldactionmodelszeroshot,li2026lingbotva}. 
However, WAMs commonly rely on iterative generation with denoising-diffusion models~\cite{zhu2025uwm,kim2026cosmospolicy} or flow-matching models~\cite{li2026lingbotva,yuan2026fastwam}, often involving both actions and future visual states. Repeated forward passes through the denoising network incur substantial latency during closed-loop control, motivating acceleration that preserves task performance.

\begin{figure}[t]
    \centering
    \includegraphics[width=\linewidth]{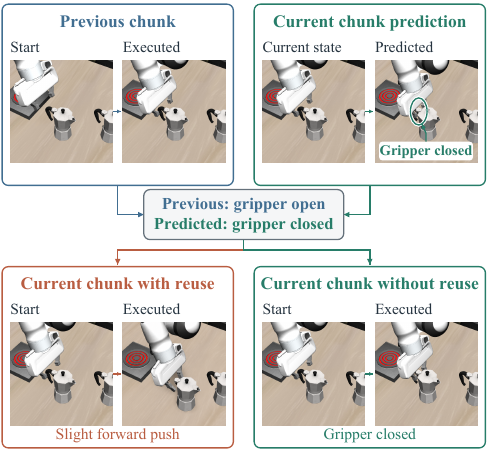}
    \caption{\textbf{Visual consistency does not imply action consistency.} Illustration of a gripper-command change across chunks. The previous chunk keeps the gripper open, while the current chunk calls for closing it. The lower-left branch illustrates how reusing the earlier behavior can push the object instead of grasping it.}
    \label{fig:outdated_behavior}
\end{figure}

Existing efforts address this cost by training more efficient architectures for future modeling and action prediction~\cite{yuan2026fastwam,ma2026fasterwam,wu2026turbowam}, or compressing multi-step generation through distillation and learned calibration~\cite{lyu2026dido,su2026onestepwam}. In contrast, training-free methods exploit redundant computation in a frozen model~\cite{ma2024deepcache,zhao2025pab}. Diffusion caching identifies similar intermediate features or predictions across denoising steps and reuses them to bypass repeated network computation~\cite{liu2025teacache,lyu2025fastercache,bu2026dicache}. However, cached features may not track the evolving denoising state, and their reuse can amplify deviations in action prediction~\cite{ji2026bac,lyu2026dido}. Transferring these methods to robot policies therefore requires assessing whether reuse accurately preserves the intended actions.

For WAM policies, consecutive inference cycles process chunks comprising action sequences~\cite{yuan2026fastwam} and future visual states~\cite{li2026lingbotva,kim2026cosmospolicy}. Existing WAM acceleration methods exploit redundancy within and across these chunks to reduce denoising network computation. Within a chunk, prediction caching exploits similar denoising updates at neighboring steps~\cite{ye2026worldactionmodelszeroshot}. Across chunks, existing methods directly reuse corresponding-step residuals~\cite{zhao2026c3ache}, or remap preceding trajectories and select retained states through observation-consistency checks~\cite{liu2026wamachine}. These approaches have advanced WAM inference acceleration by exploiting denoising redundancy. Yet visual consistency between current and reference observations does not guarantee \emph{action consistency} for upcoming behavior, which may require different motion or gripper commands. Reusing the reference information without detecting this change can propagate outdated behavior into subsequent execution. Figure~\ref{fig:outdated_behavior} illustrates this mismatch: carrying an open-gripper behavior into a chunk that requires closing can push the object rather than grasp it, despite little change in the scene. This motivates identifying cross-chunk redundancy through \emph{action intent}, the upcoming behavior predicted from the latest input, especially without reliable overlapping predictions. Effective reuse further requires adapting cached computation to the current input, since compatible intent does not imply identical denoising results. Our central question is: \emph{how can action consistency guide redundancy identification, and how should cached computation be corrected for cross-chunk reuse while preserving action quality?}

To address this question, we propose \name{}, a training-free framework for action-consistent caching. A coarse-to-fine gate identifies reusable computation by assessing recent motion and visual consistency, then checking predicted action intent against a cached reference chunk. For accepted chunks, differences between current and reference first-step predictions guide correction of cached states to the current input, providing a later denoising entry that bypasses the redundant prefix. Subsequent inference refines this estimate while reusing stable predictions at neighboring steps. This within-chunk reuse addresses denoising redundancy independently of cross-chunk acceptance. Our contributions are threefold:
\begin{itemize}
    \item We identify action consistency, including predicted upcoming behavior, as a criterion for exploiting cross-chunk denoising redundancy.
    \item We develop \name{}, a training-free framework that combines prediction-guided correction with complementary reuse across and within chunks.
    \item Experiments on three WAM families across LIBERO and RoboTwin demonstrate $1.41$--$5.44\times$ GPU inference speedup with comparable task success, alongside $1.46\times$ end-to-end speedup on the Franka Research 3 robot.
\end{itemize}

\begin{figure*}[!t]
    \centering
    \includegraphics[width=\textwidth]{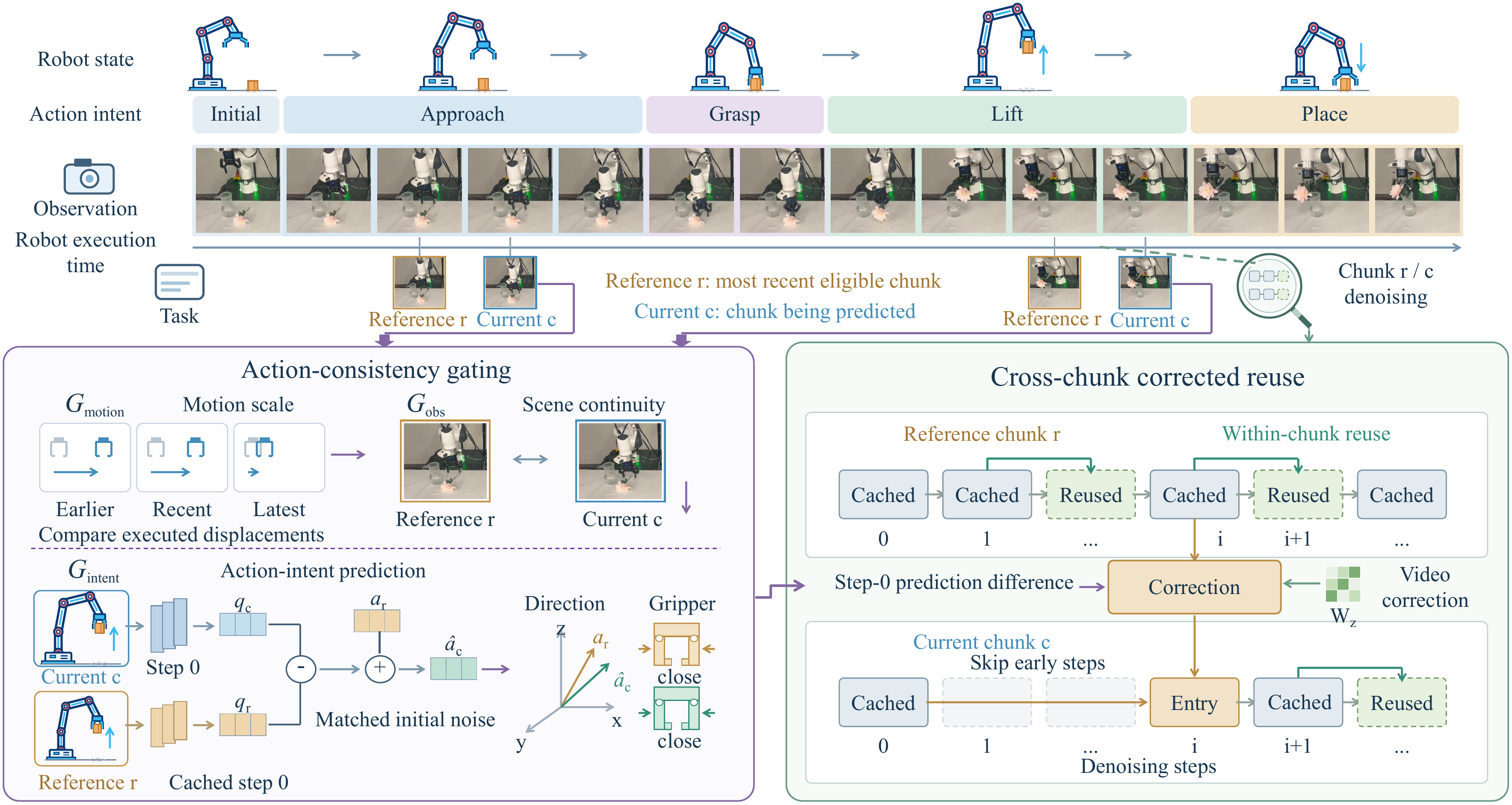}
    \caption{\textbf{Overview of \name{}.}
    Top: observations and robot states evolve as successive action chunks are executed.
    Bottom left: recent motion, visual consistency, and predicted upcoming behavior jointly assess action consistency between the current chunk $c$ and reference chunk $r$.
    Bottom right: first-step response differences correct cached states
    at entry $i$, bypassing redundant early denoising steps.
    Spatial weights $W_z$ modulate video correction, while within-chunk
    prediction reuse reduces computation during subsequent refinement.}
    \label{fig:method_overview}
\end{figure*}

\section{Related Work}
\label{sec:related_work}

\noindent\textbf{World-action models.}
DyWA~\cite{lyu2025dywa} couples action learning with future-state prediction for dynamics-adaptive manipulation. Motus~\cite{bi2026motus} unifies understanding, video generation, and action prediction, while Cosmos Policy~\cite{kim2026cosmospolicy} jointly predicts actions, future observations, and values with a pretrained video diffusion model. LingBot-VA~\cite{li2026lingbotva} autoregressively alternates visual-state prediction and action generation with observed feedback. DreamZero~\cite{ye2026worldactionmodelszeroshot} extends joint video--action modeling to zero-shot policies. More recently, $\tau_0$-WM~\cite{zhou2026tau0wm} incorporates future simulation and task-progress evaluation, and OpenWAM~\cite{wang2026openwam} provides modular WAM pretraining. Despite architectural differences, repeated denoising remains a shared bottleneck.

\noindent\textbf{Acceleration for world-action models.}
\emph{Training-based methods} reduce inference cost through architectural changes, as in FastWAM~\cite{yuan2026fastwam}, Faster-WAM~\cite{ma2026fasterwam}, and TurboWAM~\cite{wu2026turbowam}, or shorten denoising through distillation and learned calibration, as in DIDO~\cite{lyu2026dido} and 1Step-WAM~\cite{su2026onestepwam}. \emph{Training-free methods} instead exploit denoising redundancy: DiT Caching~\cite{ye2026worldactionmodelszeroshot} reuses predictions within chunks, C$^3$ache~\cite{zhao2026c3ache} transfers corresponding-step residuals across chunks, and WAMachine~\cite{liu2026wamachine} coordinates trajectory and intermediate-state reuse with observation feedback. RealtimeWAM~\cite{liu2026realtimewam} further combines parallel execution with adaptive feature and residual reuse. Our focus is identifying cross-chunk redundancy that preserves upcoming action consistency. \name{} complements motion and visual checks with predicted action intent, then corrects accepted cached computation for the current input, alongside within-chunk prediction reuse.

\section{Background and Motivation}
\label{sec:preliminaries}

\noindent\textbf{World-Action Model.}
A world-action model (WAM) is a generative robot policy that models future actions together with their visual consequences, conditioned on the current scene and task. At inference cycle $c$, the input is $\xi_c=(o_c,p_c,\ell)$, comprising visual observations, the robot state, and a language instruction. The model generates $(A_c,V_c)\sim p_\theta(\cdot\mid\xi_c)$, where $A_c=(a_{c,1},\ldots,a_{c,H})$ is a future action sequence of horizon $H$ and $V_c$ denotes the predicted future video. Each consecutive inference cycle produces one \emph{chunk}, $(A_c,V_c)$, $V_c$ is omitted for action-only policies. After an execution window, updated observations and robot state condition the next chunk.

\noindent\textbf{Iterative Denoising.}
Let $z_c^i$ denote the latent state at denoising step $i$, containing the action latent $z_{c,a}^i$ and, for models that generate video, the video latent $z_{c,v}^i$. Starting from noise $z_c^0$, the model progressively refines this state along decreasing noise levels $\sigma_0>\cdots>\sigma_K$. The denoising network $f_\theta$, with frozen parameters $\theta$, predicts the velocity $v_c^i=f_\theta(z_c^i,\sigma_i;\xi_c)$ from the current latent and noise level, conditioned on $\xi_c$. A first-order update is
\begin{equation}
 z_c^{i+1}=z_c^i+(\sigma_{i+1}-\sigma_i)v_c^i.
 \label{eq:method_solver}
\end{equation}
The final latent is decoded into $A_c$ and, where applicable, $V_c$. Equation~\ref{eq:method_solver} provides a unified description of denoising updates. In practice, \name{} retains each backbone's original sampling algorithm.

\noindent\textbf{Computation redundancy.}
\emph{Within-chunk redundancy} comes from predicting similar denoising updates at neighboring steps. Across adjacent chunks, similar action sequences suggest \emph{cross-chunk redundancy}, motivating reuse of prior computation to skip early denoising steps. The reference chunks come from a previously processed chunk whose action predictions and intermediate denoising states are cached for potential reuse. We index the current chunk by $c$ and the reference chunk by $r<c$. Beyond visual consistency, we assess predicted action intent to determine whether the upcoming actions remain compatible with this reference. A cache-age limit $A_{\max}$ bounds the number of inference cycles since the reference, $c-r\le A_{\max}$, with older references refreshed before further reuse.

\section{Methodology}
\label{sec:method}
Figure~\ref{fig:method_overview} summarizes \name{}, a training-free framework that combines cross-chunk action-consistent caching for skipping redundant early denoising steps with within-chunk similar velocity prediction reuse. A coarse-to-fine gate assesses action consistency to select reusable references. First-step prediction differences then adapt cached latents and network features to the current input, providing a later denoising entry. Subsequent steps refine this estimate while reusing stable denoising predictions.

\subsection{Action-Consistent Cross-chunk Gating}
\label{sec:intent_continuity}

To identify cross-chunk computation that can be skipped, \name{} assesses action consistency from coarse execution cues to predicted upcoming behavior. Motion-scale screening identifies candidate reuse opportunities, visual comparison checks their relevance to the current scene, and action-intent prediction detects finer behavioral conflicts with the reference.

\noindent\textbf{Motion-scale screening.}
Robotic manipulation ranges from sustained approach and transport to precision-sensitive alignment and placement. \name{} takes changes in motion magnitude as a coarse cue to distinguish potentially redundant computation from updates needed for finer control. With $d_j$ denoting net end-effector displacement over an inference cycle, the motion gate is
\begin{equation}
 G_{\mathrm{motion}}=\mathbf{1}\!\left[\frac{d_{c-1}+\varepsilon}{\min(d_{c-2},d_{c-3})+\varepsilon}\ge1\right].
 \label{eq:method_regime}
\end{equation}
Here $\mathbf{1}[\cdot]$ is the indicator function and $\varepsilon>0$ stabilizes the ratio. A contraction from recent motion suggests finer adjustment and fails this screen.

\noindent\textbf{Visual-consistency comparison.}
Motion contraction alone does not establish whether cached computation remains relevant to the current scene. For current and reference observations $o_c$ and $o_r$, we define
\begin{equation}
 G_{\mathrm{obs}}=\mathbf{1}\!\left[
 \max\left\{\frac{\operatorname{mean}|o_c-o_r|}
 {\tau\,\operatorname{mean}|o_r|}\right\}\le1
 \right].
 \label{eq:method_visual_gate}
\end{equation}
The maximum is taken over global and local comparisons of the normalized observations. Each mean is evaluated over the full observation and the local regions with the largest changes, with $\tau$ denoting the corresponding tolerance. Requiring visual comparisons to pass prevents a stable background from masking changes near the interaction.

\noindent\textbf{Predictive action-intent checking.}
Recent motion and visual consistency provide coarse redundancy evidence for reuse, but can miss changes in the upcoming action. We therefore assess \emph{action intent}, to detect pronounced reversals in the predicted motion direction and changes in gripper state before accepting reference reuse.

To assess upcoming action intent at low cost, we extrapolate
the initial denoising velocity
$v_c^0=f_\theta(\epsilon,\sigma_0;\xi_c)$ from noise level
$\sigma_0$ to zero, using initial noise $\epsilon$ shared
with the reference. The native action decoder $\mathcal D_a$
maps this clean-latent estimate to a coarse prediction of
future actions:
\begin{equation}
 q_c=\mathcal{D}_a\!\left(\epsilon-\sigma_0 v_c^0\right).
 \label{eq:method_probe}
\end{equation}
This requires one network forward pass and no additional training. We further capture changes in predicted behavior by obtaining
$q_r$ analogously from the cached reference prediction $v_r^0$. The operators $\mathcal P_c$ and $\mathcal P_r$ select
overlapping time intervals, or matching relative positions
when no overlap exists. Using the difference between the two coarse predictions
to estimate the change in upcoming actions, \name{}
constructs the proposal:
\begin{equation}
 \widehat a_c=a_r+\mathcal P_cq_c-\mathcal P_rq_r.
 \label{eq:method_intent}
\end{equation}
The intent gate compares $\widehat a_c$ with $a_r$ and accepts
reuse when $\operatorname{Conflict}(a_r,\widehat a_c)$ is false.
This conflict indicator detects opposing mean translational
commands, including vertical reversals obscured by similar
horizontal motion, or persistent disagreement in gripper
opening and closing over the selected action positions.

For an available reference, the predictive intent check and the complete action-consistency gate are
\begin{equation}
 \begin{aligned}
 G_{\mathrm{intent}}&=\mathbf{1}\!\left[\operatorname{Conflict}(a_r,\widehat a_c)=0\right],\\
 G_c&=G_{\mathrm{motion}}\land G_{\mathrm{obs}}\land G_{\mathrm{intent}}.
 \end{aligned}
 \label{eq:method_joint_gate}
\end{equation}
Cross-chunk reuse is enabled only when all three checks pass, giving $G_c=1$.

\subsection{Prediction-Guided Prefix Skipping}
\label{sec:prefix_correction}

For chunks that pass the gate, action-consistency does
not ensure that cached denoising states remain appropriate
for current condition. \name{} therefore adapts these
states through correction before resuming sampling at step
$i$, termed the entry step, bypassing the early denoising prefix.

\noindent\textbf{Action-latent correction.}
The first-step velocity difference, $v_c^0-v_r^0$, estimates
the change in denoising direction between the current and
reference chunks. We extrapolate this difference to the
entry noise level $\sigma_i$, with $\alpha\in(0,1]$
controlling the correction magnitude. This yields
\begin{equation}
 \widetilde z_{c,a}^i=z_{r,a}^i+\alpha(\sigma_i-\sigma_0)(v_{c,a}^0-v_{r,a}^0).
 \label{eq:method_entry}
\end{equation}
This estimate replaces the repeated early denoising computation, allowing steps $1,\ldots,i-1$ to be skipped.

\noindent\textbf{Spatially weighted video correction.}
For visual cross-chunk adaptation, we additionally introduce
a spatial weighting map $W_z$, derived by comparing
conditioning-related attention maps and visual features
between the current and reference chunks. It emphasizes
changes in task-relevant regions while retaining a small
correction elsewhere. For a separate video branch, denoted
by $v$, we obtain
\begin{equation}
 \widetilde z_{c,v}^i=z_{r,v}^i+(\sigma_i-\sigma_0)W_z\odot(v_{c,v}^0-v_{r,v}^0),
 \label{eq:method_spatial}
\end{equation}
where $\odot$ denotes element-wise multiplication. Together, these corrections adapt the cached action and video latents to the current input, providing an entry estimate that bypasses the redundant denoising prefix.

\begin{table*}[t]
\centering
\caption{\textbf{Inference efficiency and task success on RoboTwin 2.0.} Bold and underlining mark the best and second-best results per backbone. $^{*}$VLA-ADP prunes visual-conditioning tokens with the native denoising schedule.}
\label{tab:main_results}
\fontsize{8}{9.5}\selectfont
\setlength{\tabcolsep}{2pt}
\renewcommand{\arraystretch}{1.12}
\begin{tabular*}{\textwidth}{@{\extracolsep{\fill}}lrrrrrrrrrrrr@{}}
\toprule
& \multicolumn{4}{c}{LingBot-VA} & \multicolumn{4}{c}{FastWAM} & \multicolumn{4}{c}{OpenWAM} \\
\cmidrule(lr){2-5}\cmidrule(lr){6-9}\cmidrule(l){10-13}
Method & FLOPs $\downarrow$ & Latency $\downarrow$ & Speedup $\uparrow$ & SR $\uparrow$ & FLOPs $\downarrow$ & Latency $\downarrow$ & Speedup $\uparrow$ & SR $\uparrow$ & FLOPs $\downarrow$ & Latency $\downarrow$ & Speedup $\uparrow$ & SR $\uparrow$ \\
\midrule
Native & 237.94 & 2424.65 & 1.000$\times$ & \underline{86.30}  & 4.04 & 85.04 & 1.000$\times$ & 83.45  & 115.52 & 1036.00 & 1.000$\times$ & 88.00  \\

VLA-ADP$^{*}$ & 236.56 & 2396.93 & 1.012$\times$ & 86.20  & 3.78 & 79.40 & 1.071$\times$ & 80.90  & 107.58 & 1025.83 & 1.010$\times$ & 87.50  \\

C$^3$ache & 230.05 & 2246.98 & 1.079$\times$ & 84.80  & 3.84 & 76.19 & 1.116$\times$ & 81.00  & 104.99 & 1016.65 & 1.019$\times$ & 86.30  \\

DiT Caching & \underline{57.15} & \underline{546.08} & \underline{4.440}$\times$ & 85.60  & 3.41 & 62.85 & 1.353$\times$ & \underline{83.90}  & \underline{35.81} & \underline{475.30} & \underline{2.180}$\times$ & \underline{88.50}  \\

WAMachine & 125.46 & 1202.94 & 2.016$\times$ & 80.75  & \textbf{3.31} & \textbf{58.94} & \textbf{1.443}$\times$ & 75.75  & 70.98 & 858.59 & 1.207$\times$ & 85.70  \\

\rowcolor{methodrowblue}
\name{} & \textbf{54.07} & \textbf{496.34} & \textbf{4.885}$\times$ & \textbf{86.35}  & \underline{3.39} & \underline{59.34} & \underline{1.433}$\times$ & \textbf{83.95}  & \textbf{33.98} & \textbf{409.64} & \textbf{2.529}$\times$ & \textbf{88.55}  \\
\bottomrule
\end{tabular*}
\end{table*}

\subsection{Within-Chunk Prediction Reuse}
\label{sec:reuse_refinement}

Within-chunk redundancy concerns similar denoising predictions under the same conditioning, independently of whether a reference chunk is reusable~\cite{ye2026worldactionmodelszeroshot}. \name{} addresses it by sharing a recently computed velocity across nearby steps while continuing the latent updates.

\noindent\textbf{Adaptive prediction reuse.}
For the current step $i$ in chunk $c$, let $j<k<i$ denote the two most recent steps with a denoising-network forward pass. Their predicted velocities, $v_c^j$ and $v_c^k$, measure the stability of the denoising direction:
\begin{equation}
 \gamma_c=\frac{\langle v_c^j,v_c^k\rangle}
 {\|v_c^j\|_2\|v_c^k\|_2+\varepsilon}.
 \label{eq:method_velocity_similarity}
\end{equation}
This cosine similarity determines a bounded number $n_c$ of subsequent steps that can share the latest prediction $v_c^k$, with higher similarity permitting a longer reuse interval. The prediction supplied to the sampler is
\begin{equation}
 \widetilde v_c^i=
 \begin{cases}
 v_c^k, & 1\le i-k\le n_c,\\
 f_\theta(z_c^i,\sigma_i;\xi_c), & i-k>n_c.
 \end{cases}
 \label{eq:method_velocity_cache}
\end{equation}
The native sampler updates the latent at every step using $\widetilde v_c^i$. After $n_c$ reused predictions, a new network forward pass refreshes the velocity, and the latest two computed predictions determine the next reuse interval.
This rule reduces repeated network computation while preserving successive denoising updates. It applies to all chunks, over the remaining schedule after prefix skipping or the full schedule when cross-chunk reuse is rejected.

\noindent\textbf{Prediction at the corrected entry.}
Let \(\mathcal S_r\) contain the steps at which the denoising network was executed for the reference chunk, and \(\mathcal J\) denote the candidate entry steps. We select the earliest candidate with cached computation:
\begin{equation}
 i=\min(\mathcal{S}_r\cap\mathcal{J}).
 \label{eq:method_entry_selection}
\end{equation}
The reference latent and cached residual are taken from the selected step $i$. Following the correction principle in Eq.~\ref{eq:method_entry}, we adjust the cached residual using the current--reference difference in step-0 feature updates. Adding this corrected residual to the current entry features and applying the native output head yields $\widetilde v_{c\leftarrow r}^i$ without a full denoising forward pass. The latent update is
\begin{equation}
 z_c^{i+1}=\widetilde z_c^i+(\sigma_{i+1}-\sigma_i)\widetilde v_{c\leftarrow r}^i.
 \label{eq:method_corrected_entry_update}
\end{equation}
At $i+1$, we recompute the prediction with current conditioning and resume within-chunk reuse.

\section{Experiments}
\label{sec:experiments}

\subsection{Experimental Setup}
\label{sec:experimental_setup}

\noindent\textbf{Benchmarks and backbones.}
We evaluate three representative WAM architectures (LingBot-VA~\cite{li2026lingbotva} autoregressively alternates video and action generation, FastWAM~\cite{yuan2026fastwam} denoises actions conditioned on cached features without generating future video, and the Cosmos family couples video and action denoising) on two diverse robot-manipulation benchmarks, LIBERO~\cite{liu2023libero} and RoboTwin~2.0~\cite{chen2025robotwin2}. The Cosmos implementations are Cosmos Policy~\cite{kim2026cosmospolicy} on LIBERO and OpenWAM (based on Cosmos Predict2.5-2B and ActionDiT)~\cite{wang2026openwam} on RoboTwin. Each LIBERO suite contains 10 tasks with 50 initial states each. RoboTwin covers 50 tasks with 20 Clean and 20 Randomized trials each. Main comparisons cover LIBERO-Long and RoboTwin, with additional Native comparisons on Spatial, Object, and Goal.

\noindent\textbf{Baselines.}
We compare Native full inference with four acceleration baselines: DiT-Caching~\cite{ye2026worldactionmodelszeroshot} for within-chunk prediction reuse, VLA-ADP~\cite{pei2025vlaadp} for action-aware visual-token pruning, C$^3$ache~\cite{zhao2026c3ache} and WAMachine~\cite{liu2026wamachine}  for cross-chunk residual reuse. Methods share weights and evaluation scenes, with acceleration settings frozen. Experiments run on an RTX PRO 6000. Adaptation and implementation details appear in the separate supplementary material.

\noindent\textbf{Metrics.}
FLOPs (TFLOPs/chunk) and Latency (ms/chunk) cover model inference computation from input encoding to action generation, averaged over all predicted chunks. Speedup is Native Latency divided by method Latency under same hardware and timing scope. Timing covers 500 LIBERO-Long episodes and 200 shared RoboTwin scenes (two Clean and two Randomized per task). SR measures task success.

\begin{table}[t]
\centering
\caption{\textbf{Inference efficiency and task success on LIBERO-Long.} Highlighting and $^{*}$VLA-ADP adaptation follow Table~\ref{tab:main_results}.}
\label{tab:libero_long}
\fontsize{9}{10}\selectfont
\setlength{\tabcolsep}{2pt}
\renewcommand{\arraystretch}{1.04}
\begin{tabular*}{\columnwidth}{@{\extracolsep{\fill}}lrrrr@{}}
\toprule
Method & FLOPs $\downarrow$ & Latency $\downarrow$ & Speedup $\uparrow$ & SR $\uparrow$ \\
\midrule
\multicolumn{5}{l}{\textit{LingBot-VA}} \\
Native & 158.29 & 2083.38 & 1.000$\times$ & \textbf{96.60} \\
VLA-ADP$^{*}$ & 157.99 & 1921.26 & 1.084$\times$ & 95.00 \\
C$^3$ache & 155.84 & 1908.01 & 1.092$\times$ & 94.60 \\
DiT-Caching & \underline{49.07} & \underline{437.74} & \underline{4.759}$\times$ & 96.00 \\
WAMachine & 91.70 & 712.35 & 2.925$\times$ & \textbf{96.60} \\
\rowcolor{methodrowblue}
\name{} & \textbf{46.59} & \textbf{383.00} & \textbf{5.440}$\times$ & \underline{96.20} \\
\midrule
\multicolumn{5}{l}{\textit{FastWAM}} \\
Native & 3.74 & 73.78 & 1.000$\times$ & 93.40 \\
VLA-ADP$^{*}$ & 3.44 & 60.09 & 1.228$\times$ & 84.80 \\
C$^3$ache & 3.53 & 65.22 & 1.131$\times$ & \textbf{94.80} \\
DiT-Caching & 3.10 & 52.81 & 1.397$\times$ & \underline{94.20} \\
WAMachine & \textbf{2.99} & \textbf{45.12} & \textbf{1.635}$\times$ & 92.40 \\
\rowcolor{methodrowblue}
\name{} & \underline{3.07} & \underline{52.23} & \underline{1.413}$\times$ & \underline{94.20} \\
\midrule
\multicolumn{5}{l}{\textit{Cosmos Policy}} \\
Native & 169.89 & 1475.86 & 1.000$\times$ & 97.00 \\
VLA-ADP$^{*}$ & 168.51 & 1428.86 & 1.033$\times$ & 97.00 \\
C$^3$ache & 156.57 & 1388.59 & 1.063$\times$ & 96.20 \\
DiT-Caching & \underline{74.57} & \underline{669.75} & \underline{2.204}$\times$ & 96.60 \\
WAMachine & 92.34 & 799.93 & 1.845$\times$ & \textbf{97.60} \\
\rowcolor{methodrowblue}
\name{} & \textbf{70.16} & \textbf{616.89} & \textbf{2.392}$\times$ & \underline{97.40} \\
\bottomrule
\end{tabular*}
\vspace{-4pt}
\end{table}

\begin{table}[t]
\centering
\caption{\textbf{Task success across LIBERO suites.} SR (\%) for \name{} and Native across four suites, with their mean (Avg.).}
\label{tab:additional_suites}
\fontsize{9}{10}\selectfont
\setlength{\tabcolsep}{2pt}
\renewcommand{\arraystretch}{1.0}
\setlength{\aboverulesep}{1.2pt}
\setlength{\belowrulesep}{1.6pt}
\begin{tabular*}{\columnwidth}{@{\extracolsep{\fill}}lrrrrr@{}}
\toprule
Method & Long & Spatial & Object & Goal & Avg. \\
\midrule
\multicolumn{6}{l}{\textit{LingBot-VA}} \\
Native & 96.60 & 98.40 & 99.40 & 96.20 & 97.65 \\
\rowcolor{methodrowblue}
\name{} & 96.20 & 98.00 & 99.20 & 95.60 & 97.25 \\
\midrule
\multicolumn{6}{l}{\textit{FastWAM}} \\
Native & 93.40 & 98.20 & 99.60 & 97.00 & 97.05 \\
\rowcolor{methodrowblue}
\name{} & 94.20 & 96.60 & 99.60 & 96.80 & 96.80 \\
\midrule
\multicolumn{6}{l}{\textit{Cosmos Policy}} \\
Native & 97.00 & 97.00 & 99.60 & 96.40 & 97.50 \\
\rowcolor{methodrowblue}
\name{} & 97.40 & 97.80 & 100.00 & 98.80 & 98.50 \\
\bottomrule
\end{tabular*}
\end{table}

\vspace{-4pt}
\subsection{Analysis of Simulation Results}
\label{sec:simulation_results}

\noindent\textbf{Efficiency and task success.}
Tables~\ref{tab:main_results} and~\ref{tab:libero_long} show that \name{} achieves $1.413\times$--$5.440\times$ inference speedup, with the lowest FLOPs and Latency on LingBot-VA and the Cosmos family. It attains the highest RoboTwin SR across all three backbones and remains within one percentage point of Native on LIBERO-Long. These results support using stronger action-consistency evidence, including predicted action intent, to skip redundant denoising steps while retaining computation for behavioral changes. VLA-ADP's visual-token pruning provides smaller gains because it leaves much of the repeated denoising computation intact, motivating acceleration beyond visual redundancy.

\noindent\textbf{Redundancy analysis.}
DiT-Caching exploits within-chunk redundancy to sparsify denoising-network computation, outperforming C$^3$ache and WAMachine in speed on LingBot-VA and the Cosmos family. The latter methods emphasize cross-chunk reuse through cached residuals and retained trajectory states, respectively. \name{} addresses both scales: within-chunk reuse reduces repeated predictions, while action-consistent selection and correction enable skipping an appropriate early denoising prefix. It further reduces FLOPs and Latency over DiT-Caching and C$^3$ache in all six configurations. This joint improvement supports combining within-chunk and action-consistent caching  to reduce redundant computation while preserving task-relevant updates.

\begin{figure}[t]
\centering
\includegraphics[width=0.96\columnwidth]{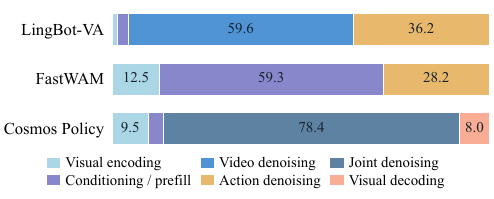}
\caption{\textbf{Backbone-dependent computation.} Estimated inference FLOPs by stage, normalized to 100\% per policy. Numbers indicate percentages.}
\label{fig:computation_bottlenecks}
\end{figure}

\noindent\textbf{Backbone-dependent acceleration.}
Figure~\ref{fig:computation_bottlenecks} explains the architectural differences in acceleration. LingBot-VA and Cosmos Policy devote most computation to denoising, making them particularly suited to \name{}'s denoising reductions. FastWAM omits future-video generation, leaving a larger share of encoding and conditioning / prefill computation that limits overall speedup from action denoising alone. Its overlapping prediction horizons also favor WAMachine's trajectory reuse, yielding higher speed, although \name{} retains substantially higher RoboTwin SR. Jointly accelerating conditioning and denoising offers further research opportunities for conditioning-heavy WAMs such as FastWAM.

\noindent\textbf{Applicability across task suites.}
Table~\ref{tab:additional_suites} extends the comparison to spatial, object, and goal variations. \name{} maintains average SR close to reproduced Native on LingBot-VA and FastWAM and achieves a higher average on Cosmos Policy. These results support action-consistent reuse across diverse manipulation demands beyond long-horizon tasks.

\subsection{Real-World Experiments}
\label{sec:real_world}

\begin{figure*}[t]
\centering
\includegraphics[width=\textwidth]{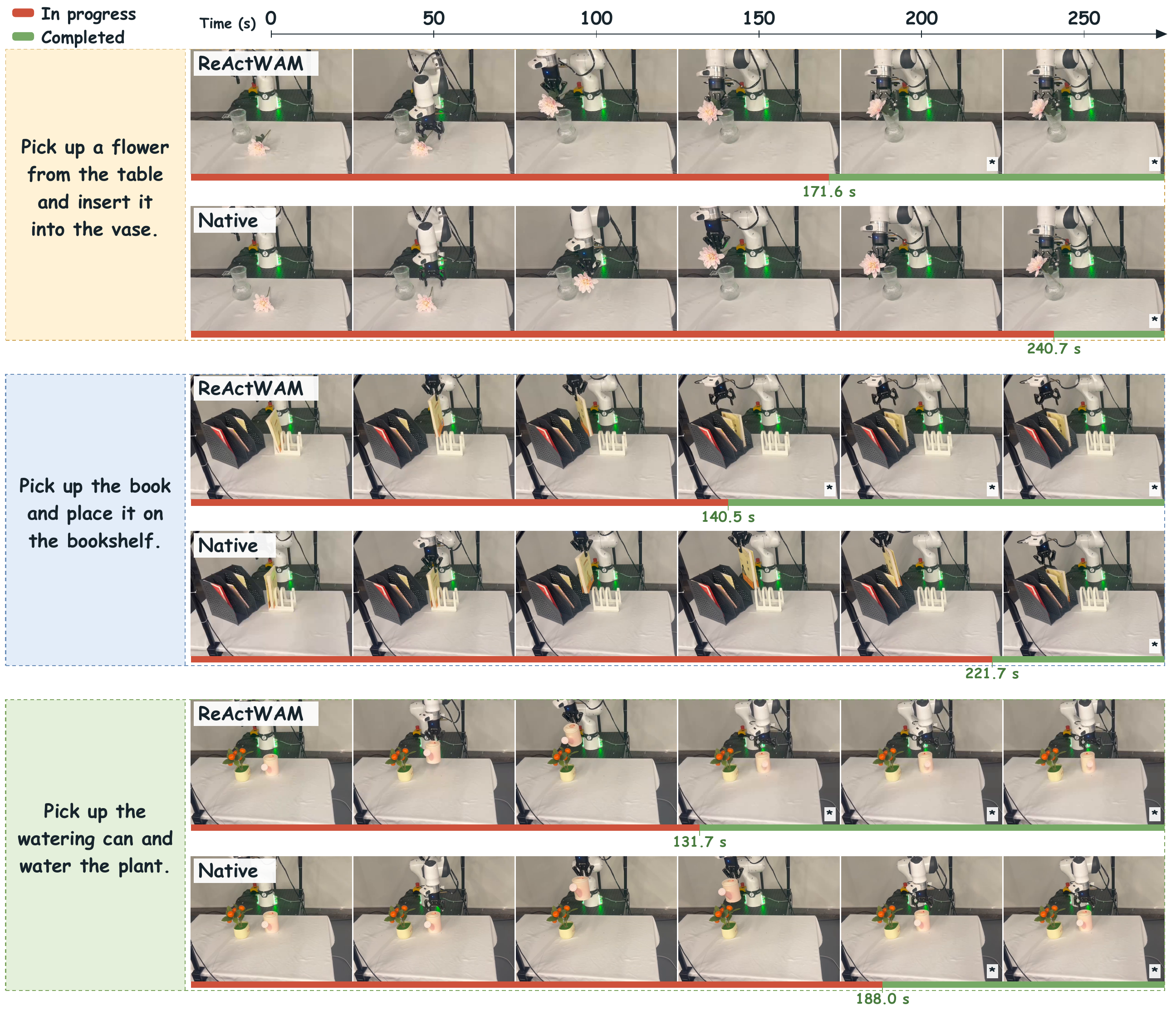}
\caption{\textbf{Time-aligned real-world executions.} Each task compares one successful rollout of \name{} (top) and Native (bottom) at six shared timestamps. Red and green indicate ongoing and completed execution. Completion times correspond to the recorded rollout endpoints, and asterisks mark final frames held for temporal alignment.}
\label{fig:real_world_keyframes}
\end{figure*}

\noindent\textbf{Experiment setup.}
We evaluate three real-world manipulation tasks on a Franka Research~3 (FR3) robot: \textit{Task 1}: pick up a flower from the table and insert it into the vase, \textit{Task 2}: pick up the book and place it on the bookshelf, \textit{Task 3}: pick up the watering can and water the plant. The policy receives RGB observations from wrist-mounted and external ZED cameras. We adopt Cosmos3-Edge-Policy-DROID with task-specific LoRA weights and a native 20-step UniPC sampling schedule, running on an NVIDIA RTX PRO 6000. We compare Native, WAMachine, DiT-Caching, and \name{} with the same task weights, with 20 trials per method and task.

\begin{table}[t]
\centering
\caption{\textbf{Real-world task success and end-to-end efficiency.} Task columns report SR (\%), Avg. reports mean SR across tasks, and Time reports mean end-to-end execution time over all test traces. Speedup is relative to Native. Bold and underlining indicate the best and second-best values, including ties.}
\label{tab:real_world}
\small
\setlength{\tabcolsep}{2.5pt}
\renewcommand{\arraystretch}{1.05}
\begin{tabular*}{\columnwidth}{@{\extracolsep{\fill}}lrrrrrr@{}}
\toprule
Method & Task 1 & Task 2 & Task 3 & Avg. & Time (s) & Speedup \\
\midrule
Native & \textbf{25.0} & \textbf{50.0} & \underline{75.0} & \textbf{50.0} & 244.8 & 1.000$\times$ \\
WAMachine & \underline{20.0} & \textbf{50.0} & 70.0 & \underline{46.7} & 186.6 & 1.312$\times$ \\
DiT-Caching & \textbf{25.0} & \underline{45.0} & \textbf{80.0} & \textbf{50.0} & \underline{178.5} & \underline{1.371}$\times$ \\
\rowcolor{methodrowblue}
\name{} & \textbf{25.0} & \underline{45.0} & \textbf{80.0} & \textbf{50.0} & \textbf{167.9} & \textbf{1.458}$\times$ \\
\bottomrule
\end{tabular*}
\end{table}

\noindent\textbf{Main result.}
Table~\ref{tab:real_world} shows that \name{} reduces mean end-to-end execution time from 244.8 to 167.9\,s ($1.458\times$), while matching Native's 50.0\% average SR. It also outperforms both acceleration baselines in execution time, with per-task SR identical to DiT-Caching. All baselines require similar numbers of chunk predictions, while representative rollouts in Figure~\ref{fig:real_world_keyframes} illustrate earlier completion with \name{}. These results support action-consistent caching as a means to improve end-to-end efficiency while preserving observed task success. Full execution videos appear in the supplementary material.

\begin{figure*}[t]
 \centering
 \begin{minipage}[t]{0.63\textwidth}
  \raggedleft
  \includegraphics[height=2.1in]{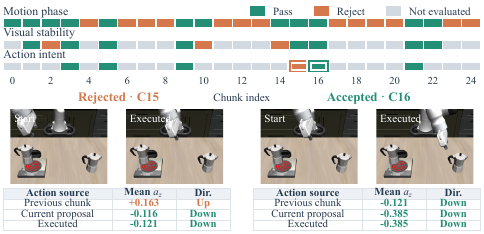}
  \par\smallskip
  {\centering\small (a) Selective reuse\par}
 \end{minipage}%
 \hspace{0.005\textwidth}%
 \begin{minipage}[t]{0.36\textwidth}
  \raggedright
  \includegraphics[width=\linewidth,height=2.1in,keepaspectratio]
    {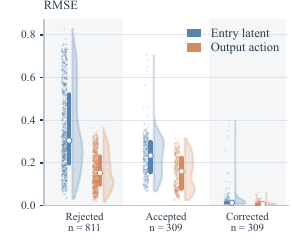}
  \par\smallskip
  {\centering\small (b) Correction fidelity\par}
 \end{minipage}
 \caption{\textbf{Selective reuse and correction fidelity.} (a) A LIBERO-Long task-8 rollout: gate decisions above, execution snapshots and action comparisons below. Previous chunk, Current proposal, and Executed denote reference actions, the low-cost action-intent prediction, and actions actually executed. Mean $a_z$ and Dir. report the mean normalized vertical command and its direction. (b) Entry-latent (blue) and final-action (orange) RMSE against full inference with matched inputs and noise over 50 task-8 episodes. Rejected/Accepted denote uncorrected reuse for chunks failing/passing the gate, and Corrected applies correction to the same accepted chunks. Dots denote chunks, white markers medians, thick bars interquartile ranges, and $n$ sample counts.}
 \label{fig:reuse_visualization}
\end{figure*}

\subsection{Ablation Study}
\label{sec:ablation}

\begin{table}[t]
\centering
\caption{\textbf{Ablations on LingBot-VA / LIBERO-Long.} Settings and metrics follow the main comparison. Bold and underlining mark the best and second-best results.}
\label{tab:ablation}
\small
\setlength{\tabcolsep}{3pt}
\renewcommand{\arraystretch}{1.05}
\begin{tabular*}{\columnwidth}{@{\extracolsep{\fill}}lrrrr@{}}
\toprule
Variant & FLOPs \(\downarrow\) & Latency \(\downarrow\) & Speedup \(\uparrow\) & SR \(\uparrow\) \\
\midrule
Native & 158.29 & 2083.38 & 1.000$\times$ & \underline{96.6} \\
\rowcolor{methodrowblue}
\name{} & 46.59 & 383.00 & 5.440$\times$ & 96.2 \\
\midrule
w/o \(G_{\mathrm{motion}}\) & 47.63 & 410.66 & 5.073$\times$ & 96.2 \\
w/o \(G_{\mathrm{obs}}\) & 45.20 & 371.53 & 5.608$\times$ & 94.8 \\
w/o \(G_{\mathrm{intent}}\) & \underline{45.03} & 369.84 & 5.633$\times$ & 94.6 \\
w/o \(G_c\) & \textbf{42.16} & \textbf{313.62} & \textbf{6.643}$\times$ & 89.2 \\
w/o \(\widetilde z_c^i\) & 45.41 & \underline{364.89} & \underline{5.710}$\times$ & 79.4 \\
w/o \(\widetilde v_c^i\) & 90.84 & 711.75 & 2.927$\times$ & \textbf{97.2} \\
\bottomrule
\end{tabular*}
\end{table}

\noindent\textbf{Settings.}
Table~\ref{tab:ablation} compares Native, \name{}, and six variants under the same cache-age, $A_{\max}=1$. Each single-gate ablation retains the other two checks, while w/o $G_c$ accepts every available chunk as reference. The variant w/o $\widetilde z_c^i$ removes both entry-latent and residual corrections, directly transferring $z_r^i$ and the cached residual at step $i$ with the current input and conditioning. The variant w/o $\widetilde v_c^i$ disables within-chunk velocity reuse while retaining cross-chunk gating, prefix skipping, and entry correction. The entry step remains unchanged, without recalibration.

\noindent\textbf{Action-consistent redundancy selection.}
Removing $G_{\mathrm{motion}}$ preserves 96.2\% SR but increases FLOPs and Latency. Without this inexpensive initial screen, reference filtering falls to $G_{\mathrm{obs}}$ and  $G_{\mathrm{intent}}$, which are more costly. Removing $G_{\mathrm{obs}}$ or $G_{\mathrm{intent}}$ instead reduces SR to 94.8\% and 94.6\%, respectively, supporting their complementary roles in checking scene relevance and upcoming behavior. Disabling all three gates yields the lowest FLOPs and Latency but reduces SR to 89.2\%. Thus, reducing cross-chunk redundancy requires action-consistent selection to preserve robust execution.

\noindent\textbf{Correction beyond direct reuse.}
Removing entry correction reduces Latency from 383.00 to 364.89\,ms but lowers SR from 96.2\% to 79.4\%. Even after passing the action-consistency gate, directly transferring the reference latent and residual cannot fully accommodate the current input. The modest correction overhead therefore supports substantially more reliable prefix skipping.

\noindent\textbf{Within-chunk redundancy reduction.}
Cross-chunk acceleration alone achieves $2.927\times$ speedup with 97.2\% SR. Adding within-chunk reuse increases speedup to $5.440\times$ while reducing SR by one percentage point, supporting the joint reduction of both forms of redundancy for a better efficiency--success trade-off.

\subsection{Visualization}
\label{sec:visualization}

Figure~\ref{fig:reuse_visualization} visualizes the coarse-to-fine gating process and representative action comparisons in a rollout, alongside the effect of correction on entry-latent and final-action errors.

\noindent\textbf{Predictive action-consistency assessment.}
In Figure~\ref{fig:reuse_visualization}(a), C15 passes motion and visual checks, but the proposal predicts a reversal and reuse is rejected. Subsequent execution without reuse confirms the predicted direction, while the remaining magnitude difference shows that adjusting an incompatible reference does not fully recover the required action. In C16, reference and proposal directions agree, allowing corrected reuse. The executed mean vertical command then matches the proposal to three decimal places. These cases show that predictive intent checking detects behavioral transitions missed by coarse checks.

\noindent\textbf{Adapting accepted computation.}
Figure~\ref{fig:reuse_visualization}(b) shows that accepted chunks have lower mean errors than rejected ones, yet substantial deviations remain under uncorrected reuse. Action consistency alone does not ensure that cached states fit the current input. Correcting these chunks reduces both entry-latent and final-action errors, indicating that entry corrections remain effective through subsequent denoising. These results support action-consistent selection and correction for reliable prefix skipping.

\section{Conclusion}
\label{sec:conclusion}

We presented \name{}, a training-free framework combining action-consistent reference selection, prediction-guided correction, and within-chunk reuse. Simulation and real-world experiments demonstrate faster inference and execution with comparable task success, supporting selective, corrected reuse for reducing denoising redundancy. Joint conditioning and denoising acceleration remains a promising research direction.

\section*{LLM Usage Statement}
We used large language models, including ChatGPT, for English-language polishing, translation, literature search, and assistance with experimental analysis. We also used general-purpose coding agents to assist with writing and debugging research code and experiment scripts. The authors reviewed AI-assisted text, code, and analyses, verified the reported results, and take full responsibility for the scientific claims and content of this paper.

{
    \small
    \bibliographystyle{ieeenat_fullname}
    \bibliography{main}
}

\end{document}